\documentclass{article}

\usepackage[square,numbers,sort&compress]{natbib}
\usepackage[preprint]{neurips_2023}

\usepackage[utf8]{inputenc} 
\usepackage[T1]{fontenc}    
\usepackage{hyperref}       
\usepackage{url}            
\usepackage{booktabs}       
\usepackage{amsfonts}       
\usepackage{amsmath}
\usepackage{nicefrac}       
\usepackage{microtype}      

\usepackage{xcolor}
\usepackage{graphicx}
\usepackage{algorithm}
\usepackage{algpseudocode}
\usepackage{multirow}
\usepackage{subcaption}
\usepackage{tikz}
\title{GAN-MPC: Training Model Predictive Controllers with Parameterized Cost Functions using Demonstrations from Non-identical Experts}

\author{%
  Returaj~Burnwal$^\dagger$\thanks{The author was a Ph.D. Student Researcher at Google Research India for the duration of this project.} \\
  Robert Bosch Centre for Data Science and AI\\
  Indian Institute of\\ Technology, Madras\\
  \texttt{returaj.burnwal$@$gmail.com} \\
   \And
   Anirban~Santara$^\dagger$ \\
   Google Research \\
   \texttt{santara$@$google.com} \\
   \AND
   Nirav~P.~Bhatt \\
   Robert Bosch Centre for Data Science and AI \\
   Indian Institute of\\ Technology, Madras \\
   \texttt{niravbhatt$@$iitm.ac.in} \\
   \And
   Balaraman~Ravindran \\
   Robert Bosch Centre for Data Science and AI \\
   Indian Institute of\\ Technology, Madras \\
   \texttt{ravindran.b$@$gmail.com} \\
   \And
   Gaurav Aggarwal \\
   Google Research \\
   \texttt{gagg$@$google.com} \\
}

\begin{document}

\maketitle
\def\thefootnote{$\dagger$}\footnotetext{These authors contributed equally to this work}\def\thefootnote{\arabic{footnote}}

\begin{abstract}
Model predictive control (MPC) is a popular approach for trajectory optimization in practical robotics applications. MPC policies can optimize trajectory parameters under kinodynamic and safety constraints and provide guarantees on safety, optimality, generalizability, interpretability, and explainability. However, some behaviors are complex and it is difficult to hand-craft an MPC objective function. A special class of MPC policies called Learnable-MPC addresses this difficulty using imitation learning from expert demonstrations. However, they require the demonstrator and the imitator agents to be identical which is hard to satisfy in many real world applications of robotics. In this paper, we address the practical problem of training Learnable-MPC policies when the demonstrator and the imitator do not share the same dynamics and their state spaces may have a partial overlap. We propose a novel approach that uses a generative adversarial network (GAN) to minimize the Jensen-Shannon divergence between the state-trajectory distributions of the demonstrator and the imitator. We evaluate our approach on a variety of simulated robotics tasks of DeepMind Control suite and demonstrate the efficacy of our approach at learning the demonstrator's behavior without having to copy their actions.

\end{abstract}

\section{Introduction}
\label{sec:introduction}
The rapid advancement in Artificial Intelligence powered by deep neural networks \citep{goodfellow2016deep} over the past decade has brought in its trail interesting developments in robotics through fast and accurate scene understanding \citep{gadre2022continuous,khan2022transformers}, ability to parse complex task specifications in natural language \citep{pmlr-v205-ichter23a,driess2023palme,stone2023openworld} and learning complex motor skills through exploration \citep{nguyen2019review,gu2017deep,haarnoja2018learning}. Still large scale deployment of robots in real-world human-centric environments remains a challenging and far fetched goal \citep{mavrogiannis2021core,alterovitz2016robot}. The most critical challenge is safety. Other important challenges are abiding by the unwritten rules of social compatibility and robustness to changes in robot dynamics and environmental conditions \cite{xiao2022learning}. Model Predictive Control (MPC) \citep{morari1988model,wang2009fast,maniatopoulos2013model,fork2021models} is a popular approach for trajectory optimization in practical robotics applications. MPC policies can optimize trajectory parameters under kinodynamic and safety constraints and provide guarantees on safety, optimality, generalizability, interpretability, and explainability. They use a model of the robot's dynamics function which defines how the robot's state changes as it interacts with the environment. However, some behaviors are complex and it is difficult to hand-craft an MPC objective function. A special class of MPC policies called \emph{Learnable-MPC} \citep{xiao2022learning} addresses this difficulty using imitation learning \citep{hussein2017imitation}. They use a parameterized objective function that can be trained from expert demonstrations. The learnable parameters also allow it to easily adapt to a wide variety of robot-environment situations. However, even the state-of-the-art Learnable-MPC formulations require the demonstrator and the imitator agents to be identical. This is an important limitation because, in most real world applications of robotics, it is not practical to assume that the dynamics of all the robots (even if they are of the same make) would be identical \citep{johansson2000state,zhao2019system}. Changes to a robot's dynamics can be caused by internal changes, such as mechanical faults \citep{verma2004real}, dropping battery charge-level \citep{hutter2017anymal}, and external changes, such as changes in the operating environment, e.g., surface friction \citep{hao2021dynamic}, or the robot's task, e.g., increased load \citep{hutter2017anymal}. Under changed dynamics, an imitator may have to implement an action, \emph{different} from the demonstrator to bring about a given state transition. In this paper, we address the practical problem of training Learnable-MPC policies when the demonstrator and the imitator do not share the same dynamics and their state spaces only have a partial overlap. This problem also has relevance in the development of personalized robotic accessibility tools for differently-abled humans \citep{qbilat2021proposal}. Our website\footnote{\url{https://sites.google.com/view/gan-mpcneurips2023}} contains videos showing our learned policies in action.

\begin{figure}[!t]
    \centering
    \includegraphics[width=0.7\textwidth]{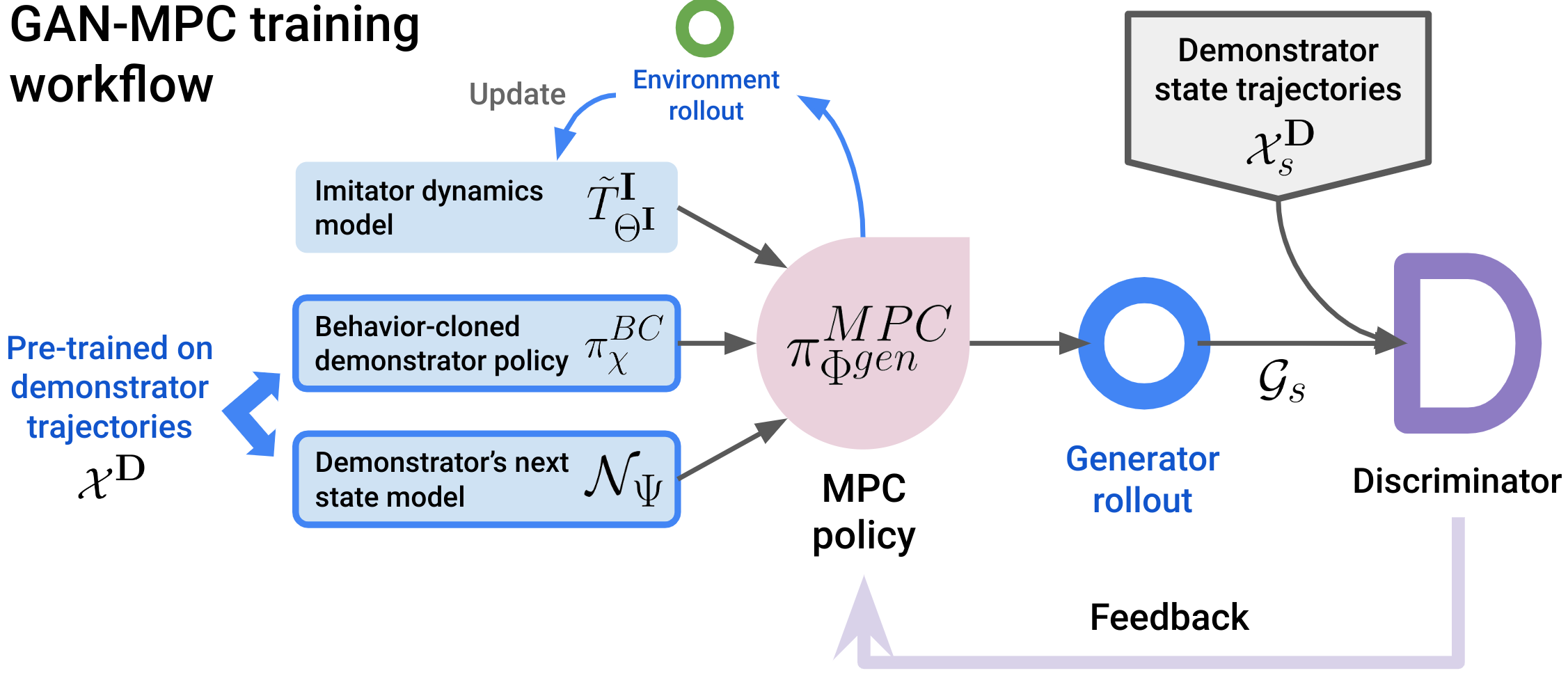}
    \caption{Description of the proposed algorithm.}
    \label{fig:gan-mpc}
\end{figure}

The problem of imitation learning from a non-identical demonstrator has been studied in the robotics community, primarily in the context of reinforcement learning (RL) \citep{liu2022plan,radosavovic2021state,liu2019state}. Deep RL has been shown to be a capable approach to learn complex skills by efficiently exploring the environment. However, these algorithms typically have high sample complexity and it is difficult to incorporate operational constraints in their objective functions. Our work in this paper is motivated by the real-world challenge of mismatched demonstrator and imitator characteristics. Therefore, we do not study RL policies and focus entirely on Learnable-MPC policies due to their practical importance.

Our proposed method uses a Generative Adversarial Network (GAN) \citep{goodfellow2020generative} to match the state-trajectory distributions of the demonstrator and the imitator by minimizing the Jensen Shannon (JS) divergence \citep{biau2020some}. The GAN consists of two networks: a generator and a discriminator. The generator is a neural network modeling the learnable cost function. This, along with the engineered cost is minimized by the imitator to produce trajectories. The discriminator is responsible for distinguishing between state trajectories from the demonstrator and the imitator. At Nash equilibrium \citep{farnia2020gans}, the state-trajectory distributions of the demonstrator and the imitator would be identical. 
Empirical evaluation on three continuous control tasks of DeepMind Control Suite \citep{tassa2018deepmind} shows that our method is effective in mimicking complex behaviors even when the dynamics of the demonstrator and the imitator are widely different. To the best of our knowledge, this is the first study on transfer learning of Learnable-MPC policies between non-identical agents.

The rest of the paper is organized as follows. Section \ref{sec:background} provides a background of our work with short introductions to essential concepts and a brief account of related works. We formally define the problem statement in Section \ref{sec:problem_definition} and present our approach in Section \ref{sec:proposed_methodology}. We experimentally evaluate our approach and compare with baselines in Section \ref{sec:experiments} and conclude the paper in Section \ref{sec:conclusion}. 

\section{Background}
\label{sec:background}
In this section, we define our notation and introduce some essential concepts that are foundational to the approach presented in this paper.

\textbf{Imitation Learning:} Imitation learning \citep{hussein2017imitation} involves two agents - demonstrator (also referred to as the ``expert'') $\mathbf{D}$ and the imitator $\mathbf{I}$. Let $\mathcal{M}^\mathbf{D} = (S^\mathbf{D}, A^\mathbf{D}, T^\mathbf{D}, \rho^\mathbf{D})$ and $\mathcal{M}^\mathbf{I} = (S^\mathbf{I}, A^\mathbf{I}, T^\mathbf{I}, \rho^\mathbf{I})$ be the Markov Decision Processes (MDPs) associated with the $\mathbf{D}$ and $\mathbf{I}$ respectively. $S^\mathbf{D}$ and $S^\mathbf{I}$ are the corresponding state spaces and $A^\mathbf{D}$ and $A^\mathbf{I}$ are the corresponding action spaces. $T^\mathbf{D}: S^\mathbf{D} \times A^\mathbf{D} \rightarrow S^\mathbf{D}$ and $T^\mathbf{I}: S^\mathbf{I} \times A^\mathbf{I} \rightarrow S^\mathbf{I}$ are the transition dynamics (interchangeably used with the shorter form - ``dynamics'') functions of $\mathbf{D}$ and $\mathbf{I}$ respectively. They predict the next state given the action taken in the current state. We use parameterised models of the dynamics functions - $T^\mathbf{D} = T^\mathbf{D}_{\Theta^\mathbf{D}}$ and $T^\mathbf{I} = T^\mathbf{I}_{\Theta^\mathbf{I}}$ where $\Theta^\mathbf{D}$ and $\Theta^\mathbf{I}$ are trainable sets of parameters. The reward functions $R^\mathbf{D}: S^\mathbf{D} \times A^\mathbf{D} \rightarrow \mathbb{R}$ and $R^\mathbf{I}: S^\mathbf{I} \times A^\mathbf{I} \rightarrow \mathbb{R}$ provide task-specific feedback to $\mathbf{D}$ and $\mathbf{I}$. These rewards are accumulated over the trajectories to measure the performance of the respective agents. $\rho^\mathbf{D}$ and $\rho^\mathbf{I}$ are the initial state distributions. $\mathbf{D}$ records a set of trajectories $\mathcal{X}^\mathbf{D}$. Trajectories, by default, refer to state-action trajectories $\tau = (s_0, a_0, s_1, a_1, s_2, a_2, \dots )$. We denote state-only trajectories as $\tau_s = (s_0, s_1, s_2, \dots )$ and the set of state-only demonstration trajectories as $\mathcal{X}^\mathbf{D}_s$. A policy is defined as a function that maps states to actions: $\pi:S\rightarrow A$. Learning a policy that mimic's the demonstrator is the ultimate aim of all imitation learning algorithms.\\

Imitation learning algorithms can be classified into two broad categories - state-action imitation learning and state-only imitation learning - based on whether the demonstrator's actions are used by the imitator \citep{hussein2017imitation}. \textbf{State-action imitation learning} falls into two categories - Behavioral Cloning \citep{bain1995framework,pomerleau1988alvinn,florence2022implicit} and Apprenticeship Learning \citep{abbeel2004apprenticeship}. In \textbf{Behavioral Cloning (BC)}, the imitator is trained to predict the demonstrator's actions given a state as input in a supervised fashion. This method is simple yet effective and does not require the imitator to interact with the environment. However, it can be brittle due to the covariate shift problem \citep{chang2021mitigating,spencer2021feedback}. Several methods have been proposed to deal with this issue including \citet{ross2011reduction} which proposes to let the imitator collect on-policy data by interacting with the environment and mix these trajectories with the demonstrator's trajectories for training.
\textbf{Apprenticeship Learning (AL)} \citep{abbeel2004apprenticeship,aghasadeghi2011maximum,ng2000algorithms,fu2017learning} involves two steps - a) learning a model of the expert's reward function by Inverse Reinforcement Learning (IRL), and b) Reinforcement Learning (RL) using the reward model from the previous step. While the performance of a BC agent is upper-bounded by the performance of the demonstrator, an AL agent does not have any such limitations. The main challenge of AL stems from the fact that IRL is an ill-defined problem. This is because, there can be multiple reward functions whose optimal policies can produce the same set of demonstrated trajectories. The maximum entropy IRL framework introduced in \citep{aghasadeghi2011maximum} addresses this issue by choosing the solution that makes minimum extra assumptions about the policy.
\textbf{State-only} or \textbf{observation-only imitation learning} algorithms cater to the case where the demonstrator's actions are not available to or usable by the imitator. Some of these algorithms work by matching the distributions of states encountered by the demonstrator and the imitator \citep{kidambi2021mobile,pmlr-v97-sun19b,zhu2020off,kostrikov2019imitation,torabi2019adversarial}. Others work by trying to infer the expert's actions using an inverse dynamics model and regularizing policy search using the predicted actions \citep{radosavovic2021state,liu2019state,edwards2019imitating,guo2019hybrid}. \citet{yang2019imitation} takes a hybrid approach which simultaneously minimizes the divergence between state-transition distributions and the disagreement between the inverse dynamics of the demonstrator and the imitator. The demonstrator's state trajectories have also been used to guide on-policy exploration through reward shaping in policy gradient based RL algorithms \citep{lee2021generalizable,lee2019efficient}. As these algorithms do not require the expert's actions, they can be used to learn motor skills by watching videos \citep{sermanet2018time,torabi2019recent} and imitate agents that do not share the same dynamics \citep{liu2022plan,radosavovic2021state,liu2019state}. Our problem setting is similar to state-only imitation learning, as differences in the transition dynamics reduce the usefulness of the expert's actions. However, we do use the expert's actions as an initial (albeit suboptimal) solution to the MPC solver, which helps the solver converge faster.

\textbf{Model Predictive Control (MPC):} Model Predictive Control \citep{morari1988model,wang2009fast,maniatopoulos2013model,fork2021models} is a closed-loop controller where the feedback received from the environment for the previous action is used to predict the next action. The closed-loop design serves to avoid divergence from the expected result, often faced by open-loop controllers where the output signal is not used for the prediction. MPC is a popular choice for trajectory planning with real-time feedback for robots operating in real-world environments that must satisfy given safety and kinodynamic constraints. Equation \ref{eqn:mpc} describes the optimization problem solved by MPC.
\vspace{-0.2in}
\begin{center}
\begin{eqnarray}
\label{eqn:mpc}
    \mathbf{a}^*_{1:H-1} = & arg\min_{\mathbf{a}_{1:H-1}} J(s_t, a_{1:H-1})\\ \nonumber
    = & arg\min_{\mathbf{a}_{1:H-1}} \sum\limits_{t=1}^{H-1} C_{stg}(s_t, a_t, t) + \gamma C_{term}(s_H)\\ \nonumber
 s.t. & \forall t, \; \; s_{t+1} = \tilde{T}(s_t, a_t), \; g(s_t, a_t) = 0, \; h(s_t, a_t) \leq 0
\end{eqnarray}
\end{center}

$H$ is the planning horizon of the MPC. $C_{stg}: S \times A \rightarrow \mathbb{R}$ is the staging cost that applies to each step of the plan and $C_{term}: S \rightarrow \mathbb{R}$ is the terminal cost that applies only to the final state. $g: S \times A \rightarrow \mathbb{R}$ and $h: S \times A \rightarrow \mathbb{R}$ are equality and inequality constraints on the solution. $\gamma$ is a hyperparameter that controls the relative weightage of the staging and the terminal costs. $\tilde{T}$ is a local model of the transition dynamics $T$ around the initial control guess. At every step of planning, the MPC plans a trajectory $\mathbf{a}^*_{1:H-1}$ of length $H$ that minimizes the objective in Equation \ref{eqn:mpc}. To address the inevitability of modeling error in the estimation of $\tilde{T}$, MPC only executes the first action $a^*_1$ and updates $\tilde{T}$ with the observed outcome. We denote an MPC policy by $\pi^{MPC}: S \rightarrow A$ where $\pi^{MPC}(s_t) = a^*_1$. This planning algorithm is repeated for every step of the agent's trajectory.

\textbf{MPC Solvers:} There are two classes of solvers used for trajectory optimization in MPC - unconstrained and constrained. Unconstrained solvers are faster but they are unable to handle the inequality constraints defined by $h(\cdot)$ in Equation \ref{eqn:mpc}. Two notable solvers in this category are Iterative Linear Quadratic Regulator (iLQR) and Differential Dynamic Programming (DDP) \citep{underactuated}. The iLQR algorithm takes as input, an ``initial control guess'' $a^g_0, a^g_1, \dots, a^g_{H-1}$. The action for time $t$, $a_t$ is obtained by minimizing the cost $J(\tilde{s}_t, a | a^g_t)$ for time $t$ defined as in Equation \ref{eqn:ilqr}.
\vspace{-0.2in}
\begin{center}
\begin{eqnarray}
\label{eqn:ilqr}
    a_t &=& arg\min_{a \in A} J(\tilde{s}_t, a | a^g_t) \\ \nonumber
    \text{where}\; \; J(\tilde{s}_t, a | a^g_t) &=& \tilde{C}_{stg}(\tilde{s}_t, a | a^g_t) + J(\tilde{s}_{t+1}, a | a^g_{t+1})
\end{eqnarray}
\end{center}
Here, $\tilde{s}_{t+1} = \tilde{T}(\tilde{s}_t, a^g_t)$, $\tilde{s}_0 = s_0$ is the observed initial state and $J(\tilde{s}_H) = C_{term}(\tilde{s}_H)$.  $\tilde{T}$ is a linear  approximation of the dynamics function and $\tilde{C}(\cdot, \cdot | a^g_t)$ is a quadratic approximation of the cost function about $(\tilde{s}_t, a^g_t)$. The solution approach of DDP is similar to iLQR with the only difference that it uses a second order approximation of the dynamics function. Usually DDP converges just as fast as iLQR and the added cost of computing the second order derivative in DDP is not worth it. This makes iLQR a more popular choice. Constrained solvers incorporate constrains into the objective function using methods like penalties, augmented Lagrangian method and the interior point method. The drawbacks of MPC arise from the requirement to handcraft the cost function and modeling the dynamics function. Also, in complex non-linear dynamical systems the quality of the solution is often a strong function of the choice of initial action.

\textbf{Learnable-MPC:} Deploying MPC in complex environments involves tedious engineering of world representations, modeling robot kinematics, hand-crafting cost functions and designing backup planners to recover from stuck situations. The robustness of such a controller is limited to the designer's anticipation of the conditions that can be encountered after deployment. It is often difficult to handcraft the MPC cost function for complex behaviors. This motivated a new class of MPC algorithms called Learnable-MPC (a term coined by \citet{xiao2022learning}) that provide parameterized cost functions. The parameters can be tuned in a data driven way to appropriately balance the different behavioral criteria \citep{shridhar1997tuning,shridhar1998tuning,garriga2010model,edwards2021automatic,yamashita2016tuning,ramasamy2019optimal}. With a goal to encode nuanced social behaviors into indoor navigation policies, \citet{xiao2022learning} presented a Learnable-MPC formulation with a parameterised terminal cost function $C_{term}(\cdot | \Phi)$. The parameters $\Phi$ are trained by minimizing the square of the L2 distance between the trajectories generated by running the controller and a set of expert demonstrations. In this paper, we extend this formulation of imitation learning of MPC policies to the case where the demonstrator and the imitator are non-identical.

\section{Problem Definition}
\label{sec:problem_definition}
Motivated by real world applications in robotics and accessibility, we study the problem of imitation learning of Learnable-MPC policies when the demonstrator and the imitator do not share the same dynamics - $T^\mathbf{D} \neq T^\mathbf{I}$. 
Our method can also be applied to settings where the state and action spaces do not overlap completely, by considering only the overlapping state and action variables.

\subsection{Challenges}
\label{sec:challenges}
MPC requires a model of the transition dynamics for planning. This is challenging in real world complex continuous control tasks with large state-action spaces. Some parts of the state-action space are difficult to reach and hence difficult to collect data from. Also, parts of the state-action space are often inaccessible due to hard kinodynamic constraints. Neural networks provide an efficient way of modeling highly non-linear functions over large state-action spaces. However, they find it hard to model the constraints and end up halucinating in the inaccessible areas, often leading to infeasible solutions. As mentioned in Section \ref{sec:background}, MPC solvers like iLQR can be highly sensitive to the ``initial control guess'' in complex non-linear dynamical systems. The challenge is to predict an $a^g_{0:H-1}$ close to the optimal solution $a^*_0$. The terminal cost $C_{term}$ is used to measure how close the agent would get to a ``target'' state at the end of the planning horizon $H$. For dynamic tasks like \texttt{Cheetah Run} the target state is different for each time step and making it difficult to calculate $C_{term}$.
\vspace{-0.1in}
\section{Proposed Methodology}
\label{sec:proposed_methodology}
 This section presents GAN-MPC, the proposed imitation learning algorithm for Learnable-MPC policies when the demonstrator $\mathbf{D}$ and the imitator $\mathbf{I}$ do not share the same dynamics. Popular Learnable-MPC formulations train the learnable parameters of the cost function by minimizing the L2-loss between the state-action trajectories of $\mathbf{D}$ and $\mathbf{I}$. This makes sense when the same action is capable of bringing about the same state transition in both the demonstrator and the imitator. In these cases, having identical transition dynamics is a necessary condition. In contrast to these cases, we deal with the situation when $\mathbf{D}$ and $\mathbf{I}$ do not share the same dynamics. Hence,  the sets of actions required to bring about a given state transition will likely be different, and the proposed algorithm in this work matches state-only trajectories $\tau_s$. Additionally, in our setting, starting from the same initial state, it may be impossible for $\mathbf{D}$ and $\mathbf{I}$ to traverse identical sequences of states. For example, a lighter $\mathbf{D}$ may inherently be able to jump higher or run faster than a heavier $\mathbf{I}$. If the height of the center-of-mass or speed is a state variable, then $\mathbf{I}$ would never be able to reach some of the states in the demonstrated trajectories. Hence, in our setting, it is important to emphasize more on achieving functionality over performance. Performance as measured by the reward function of $\mathbf{D}$ might not be appropriate for an $\mathbf{I}$ with different dynamic properties. However, $\mathbf{I}$ should learn to demonstrate behavior that is closer to the way $\mathbf{D}$ approaches the task. We achieve this by minimizing the Jensen Shannon (JS) divergence between the state-trajectory distributions of $\mathbf{D}$ and $\mathbf{I}$ within a Generative Adversarial Network (GAN) training framework.

The GAN framework involves a two-player competitive zero-sum game between two agents - a generator and a discriminator. Given a set of real data samples, the task of the discriminator is to learn an accurate binary classifier to tell apart real samples from fake ones. The task of the generator is to produce samples that are indistinguishable from real samples by the discriminator. In our setting, each sample is a state-trajectory. The generator is the Learnable-MPC policy $\pi^{MPC}(\cdot|\Phi^{gen})$ of $\mathbf{I}$ along with a model of the transition dynamics $\tilde{T}^\mathbf{I}$. $\Phi^{gen}$ is the set of learnable parameters of the terminal cost function. Given a demonstrated state-trajectory $\tau^\mathbf{D}_s = (s^\mathbf{D}_0, s^\mathbf{D}_1, s^\mathbf{D}_2, \dots) \in \mathcal{X}^\mathbf{D}_s$, a generator rollout $\tau^{\mathbf{I},g} = (s^{\mathbf{I},g}_0, a^{\mathbf{I},g}_0, s^{\mathbf{I},g}_1, a^{\mathbf{I},g}_1, s^{\mathbf{I},g}_2, a^{\mathbf{I},g}_2, \dots, s^{\mathbf{I},g}_{P-1})$ of maximum length $P$ (a hyper parameter) is created by starting from the same initial state $s^{\mathbf{I},g}_0 = s^{\mathbf{D}}_0$, solving for actions using the MPC policy $a^{\mathbf{I},g}_t = \pi^{MPC}(s^{\mathbf{I},g}_t)$ and the next state from the transition dynamics model $s^{\mathbf{I},g}_{t+1} = \tilde{T}^{\mathbf{I},g}(s^{\mathbf{I},g}_t, a^{\mathbf{I},g}_t)$. We denote the state trajectory distribution of the generator rollouts by $\mathcal{G}_s(\cdot | \Phi^{gen}, \Theta^\mathbf{I})$. The discriminator $Q(\cdot | \Phi^{disc})$ is modelled using an LSTM network with parameters $\Phi^{disc}$.

The performance of an MPC policy is strongly dependent on the accuracy of transition dynamics model $T$. As noted in Section \ref{sec:challenges} learning a model of $T^\mathbf{I}$ can be challenging in large state-action spaces. 
The dynamics function must be trained on $(s_t, a_t, s_{t+1})$ transitions collected by the agent while interacting with the environment. In order to model the function accurately in the regions of the state-action space traversed during the execution of the target task, enough data must be collected from those regions. This is not a big issue when $\mathbf{D}$ and $\mathbf{I}$ are identical as the demonstrated trajectories $\mathcal{X}^\mathbf{D}$ can be used for training $T^\mathbf{I}$. However, in our case, getting $\mathbf{I}$ to the desired regions of the state-action space can be as hard as learning the policy. We address this challenge by by pre-training $T^\mathbf{I}$ on $\mathcal{X}^\mathbf{D}$ for a small number of epochs $N^{pre}$ under the assumption that the demonstrator and the imitator dynamics have some degree of similarity. We continue to update the dynamics model in each training iteration with transitions recorded from physical interaction of $\mathbf{I}$ with the environment with $\pi^{MPC}$.

We use the popular iLQR solver in our experiments. As noted in Section \ref{sec:challenges}, the performance is a strong function of the initial control guess $a^{\mathbf{I},g}_{0:H-1}$. We again make the assumption that the demonstrator and imitator dynamics have some degree of similarity. We train a behavior cloning policy $\pi^{BC}_\chi: S^\mathbf{I} \rightarrow A^\mathbf{I}$ with parameters $\chi$ on $\mathcal{X}^\mathbf{D}$. At each iteration of iLQR, we set $a^{\mathbf{I},g}_t = \pi^{BC}(\tilde{s}^{\mathbf{I}}_t)$.

The terminal component of the MPC cost function $C_{term}$ is intended to estimate how far the agent would be from the target state at the end of the planning horizon. In dynamic tasks like \texttt{Cheetah Run}, the target state is not singular making it difficult to specify $C_{term}$. With a motivation to set as target state as somewhere the expert would be in the next time step, we train a neural network model $\mathcal{N}_\Psi:S^\mathbf{D} \rightarrow S^\mathbf{D}$ with trainable parameters $\Psi$ on $\mathcal{X}^\mathbf{D}$ to predict the next state $s^\mathbf{D}_{t+1}$ given the current state $s^\mathbf{D}_{t}$.

Our algorithm starts  by pre-training the dynamics model of the imitator on $\mathcal{D}$ for a small number of epochs $N^{pre}$. In the main training loop, in the first step, we let the imitator interact with the environment for $K$ time steps and use this data to update the dynamics model by running a small number of epochs $N^{dyn}$ of training. Next, the discriminator network is trained on $\mathcal{D}^s$ and the imitator's state trajectories. In the final step, the learnable parameters of the MPC policy and the relative weight of the engineered and learnable cost components are updated slowly. Algorithm \ref{algo:gan-mpc} presents the pseudocode. 

\begin{algorithm*}[!t]
\textbf{Input:} Set of demonstrated trajectories $\mathcal{X}^\mathbf{D}$, terminal cost weight $\gamma$, MPC planning horizon $H$, MPC training steps $N^{MPC}\in \mathbb{Z}^+$, a small integer $K \in \mathbb{Z}^+$, discriminator batch size $B\in 2\mathbb{Z}^+$, and maximum length of generator rollouts $P\in \mathbb{Z}^+$.

\textbf{Initialization:} 

\begin{enumerate}
    \item \textbf{Initialize neural network parameters} -- $\Theta^\mathbf{I}$, $\Phi^{gen}$, $\Phi^{disc}$, $\chi$ and $\Psi$ -- from a Glorot-Uniform distribution.
    \item \textbf{Pre-train imitator dynamics on demonstrator transitions}:
    \begin{equation}
        \Theta^\mathbf{I} \leftarrow arg\min_{\Theta^\mathbf{I}}  \mathop{\mathbb{E}}_{(s, a, s')\sim \mathcal{X}^\mathbf{D}}\left[ (s' - \tilde{T}^\mathbf{I}_{\Theta^\mathbf{I}}(s, a))^2 \right]
    \end{equation}

    \item \textbf{Train behavior cloning policy on demonstrator actions}:
    \begin{equation}
        \chi \leftarrow arg\min_{\chi} \mathop{\mathbb{E}}_{(s, a)\sim \mathcal{X}^\mathbf{D}}\left[ (a - \pi^{BC}_{\chi}(s))^2 \right]
    \end{equation}

    \item \textbf{Train demonstrator's next state prediction model}:
    \begin{equation}
        \Psi \leftarrow arg\min_{\Psi} \mathop{\mathbb{E}}_{(s, s')\sim \mathcal{X}^\mathbf{D}_s}\left[ (s' - \mathcal{N}_{\Psi}(s))^2 \right]
    \end{equation}

\end{enumerate}
\textbf{Main training loop:}\\
Initialize the set of imitator interactions: $\mathcal{R} \leftarrow \phi$. \\
$\Phi^{gen}[0] \leftarrow \Phi^{gen}$, $\Phi^{disc}[0] \leftarrow \Phi^{disc}$\\
Repeat for $n = 1, 2, \ldots, N^{MPC} $:
\begin{enumerate}
    \item \textbf{Imitator rollouts:} Roll out $K$ trajectories $\{\tau^\mathbf{I}_i \sim \pi^{MPC}(\cdot|\Phi^{gen}[n-1])\}_{i=1}^{K}$ by letting the imitator interact with the environment and append to $\mathcal{R}$.\\
    $\mathcal{R} \leftarrow \mathcal{R}\cup \{\tau^\mathbf{I}_i\}_{i=1}^{K}$\\
    
    \item \textbf{Update the imitator transition dynamics:} Fine tune the model on $\mathcal{R}$.
    \begin{equation}
        \Theta^\mathbf{I} \leftarrow arg\min_{\Theta^\mathbf{I}} \mathop{\mathbb{E}}_{(s, a, s')\sim \mathcal{R}}\left[ (s' - \tilde{T}^\mathbf{I}_{\Theta^\mathbf{I}}(s, a))^2 \right]
    \end{equation}

    \item \textbf{Generator rollouts: } Randomly sample $B/2$ demonstrator trajectories from $\mathcal{X}^\mathbf{D}$. Sample an equal number ($B/2$) of generator trajectories from $\mathcal{G}_s(\cdot|\Phi^{gen}[n-1], \Theta^\mathbf{I})$ using the same initial states as the demonstrator trajectories, $\pi^{MPC}_{\Phi[n-1]}$ for actions and $\tilde{T}^{\mathbf{I}}_{\Theta^{\mathbf{I}}}$ for next states as explained in Section \ref{sec:proposed_methodology}.
    \item \textbf{Discriminator update: } Use the $B$ state-trajectories from the previous step to update the discriminator network parameters $\Phi^{disc}$.
    \begin{eqnarray}
        \Phi^{disc}[n] =& arg\max_{\Phi^{disc}} \mathop{\mathbb{E}}_{\tau_s \sim \mathcal{X}^{\mathbf{D}}_s} [\log(Q(\tau_s|\Phi^{disc}[n-1]))] + \\ \nonumber
        & \mathop{\mathbb{E}}_{\tau_s \sim \mathcal{G}_s(\cdot|\Phi^{gen}[n-1], \Theta^\mathbf{I})} [\log(1 - Q(\tau_s | \Phi^{disc}[n-1])]
    \end{eqnarray}

    \item \textbf{Generator update: } Use the Polyak scheme to slowly update the generator parameters $\Phi^{gen}$ using the generator state-trajectories.
    \begin{equation}
        \Phi^{gen}[n] = arg\min_{\Phi^{gen}} \mathop{\mathbb{E}}_{\tau_s \sim \mathcal{G}_s(\cdot|\Phi^{gen}[n-1], \Theta^\mathbf{I})} [\log(1 - Q(\tau_s|\Phi^{disc}[n])]
    \end{equation}
\end{enumerate}
\caption{\label{algo:gan-mpc}Pseudocode of the proposed ``GAN-MPC'' algorithm.}
\end{algorithm*}

\section{Experiments}
\label{sec:experiments}

\begin{figure}[!t]
    \centering
    \includegraphics[width=0.9\linewidth]{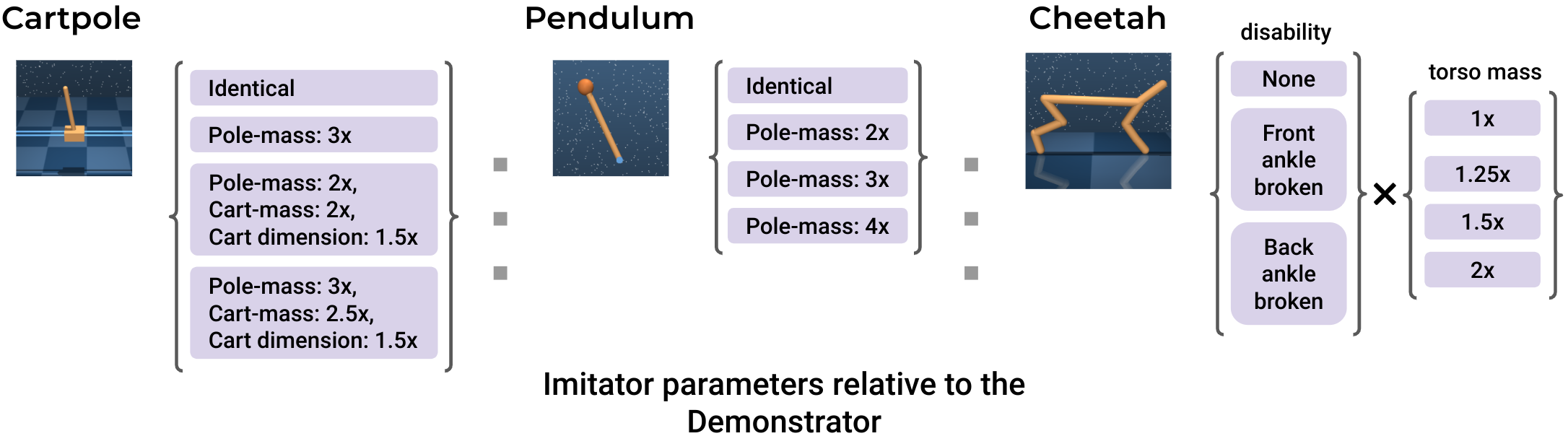}
    \caption{Physical properties of the imitators relative to the demonstrators in our experiments. We have $4$ imitators each for \texttt{Cartpole-Balance} and \texttt{Pendulum-Standup}. In case of \texttt{Cheetah-Run}, we have $12$ imitators with different levels of disability and different torso-masses as denoted by the set product ``$\times$'' in the figure.}
    \label{fig:imitators}
\end{figure}

We present an empirical study to assess the efficacy of the GAN-MPC approach presented in Section \ref{sec:proposed_methodology}. We choose three continuous control task-environments from the DeepMind Control \citep{tassa2018deepmind} suite: \texttt{CartPole-Balance}, \texttt{Pendulum-Standup} and \texttt{Cheetah-Run}. For each environment, we train an agent with default dynamics using Soft Actor Critic (SAC) \citep{haarnoja2018soft} for use as a demonstrator. SAC is a model-free RL algorithm. We train each demonstrator for 3 million episodes to achieve high rewards -- see Table \ref{tab:expert-rewards} in the supplementary material for details. We choose a set of imitator agents that have similar morphology as the demonstrators but different physical properties as described in Figure \ref{fig:imitators}. While similarity in morphology allows us to assume that the optimal action of an imitator would be close to that of the demonstrator in a given state, the differences in physical properties create differences in dynamics. We take several practical measures for stabilising the training of GAN in our framework. We use R1 regularization \citep{mescheder2018training} in the discriminator cost function and global weight clipping \citep{adler2018banach} in the Adam optimizer \citep{kingma2014adam} used for minimization of the cost function. We also update the generator with Polyak averaging \citep{polyak1992acceleration} of the parameters. The goal of our experiments is to study whether GAN-MPC can learn an expert's skills by trying to visit the same sequence of states and planning an appropriate sequence of actions, even though the imitator's actions may be different from the expert's due to differences in dynamics.

\begin{figure}
\centering
\begin{subfigure}{.25\textwidth}
\includegraphics[width=\linewidth]{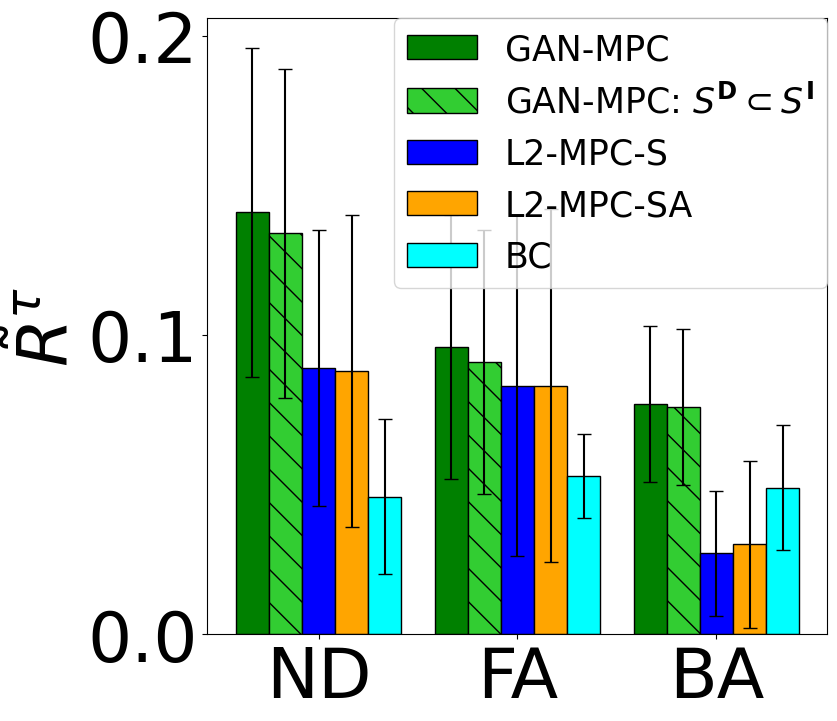}
\caption{Same torso-mass}
\end{subfigure}%
\begin{subfigure}{.25\textwidth}
\includegraphics[width=\linewidth]{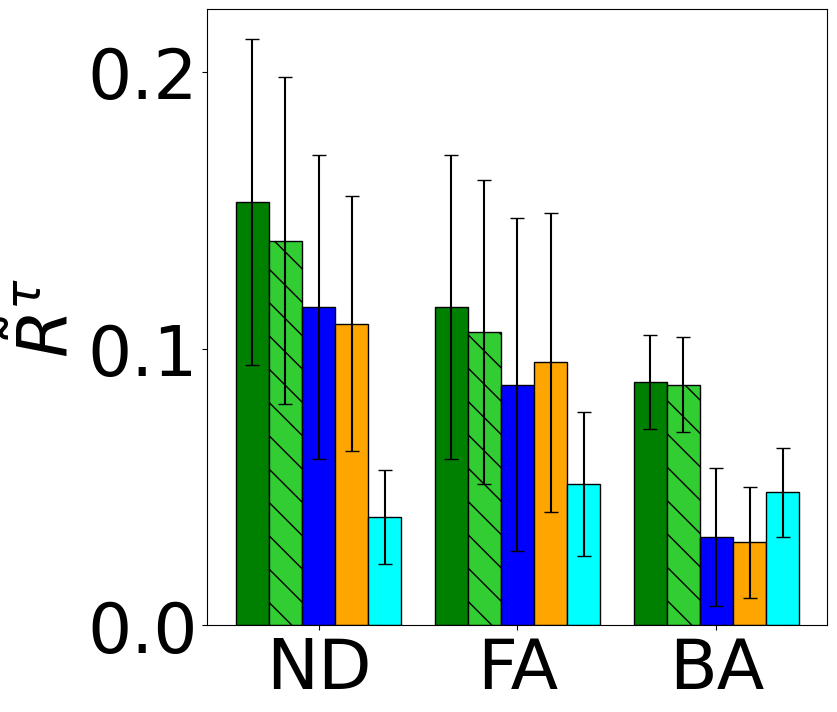}
\caption{$1.25\times$ torso-mass}
\end{subfigure}%
\begin{subfigure}{.25\textwidth}
\includegraphics[width=\linewidth]{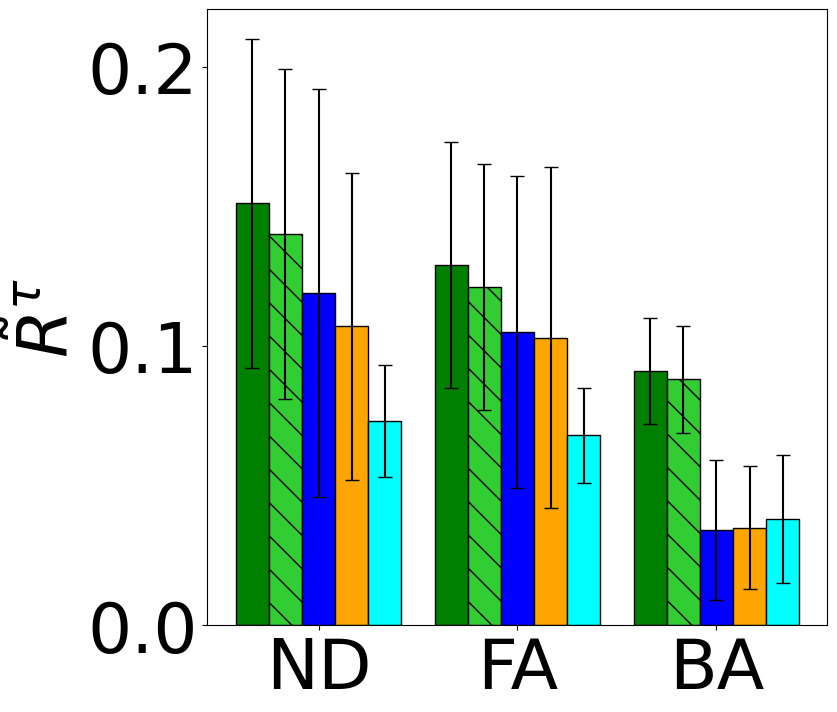}
\caption{$1.5\times$ torso-mass}
\end{subfigure}%
\begin{subfigure}{.25\textwidth}
\includegraphics[width=\linewidth]{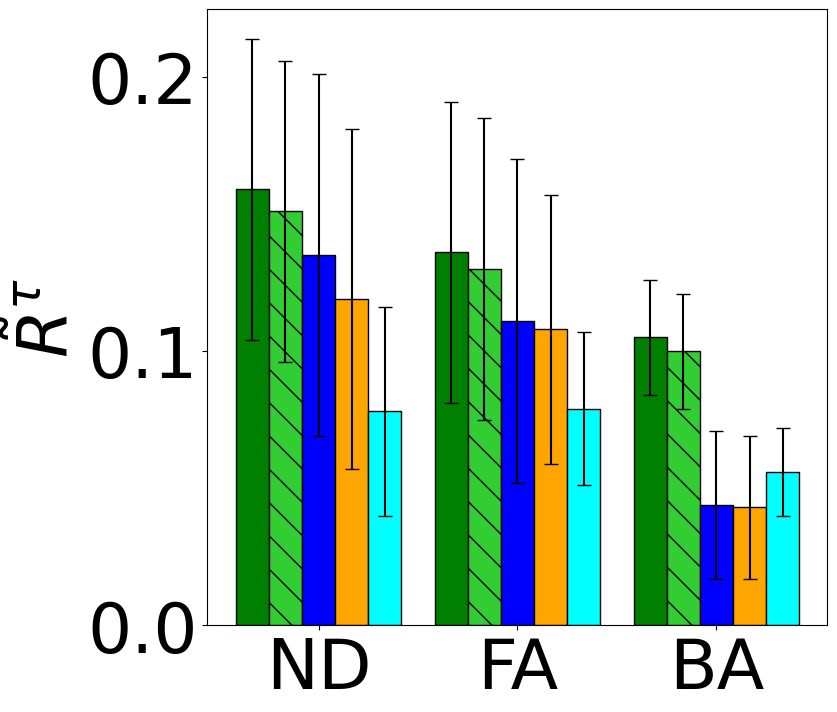}
\caption{$2\times$ torso-mass}
\end{subfigure}
\caption{Results of the \texttt{Cheetah-Run} experiment. The captions of the sub-figures mention ``torso-mass'' of the imitators relative to the demonstrator. As described in Section \ref{sec:experiments} and Figure \ref{fig:imitators}, we have three categories of imitators in terms of disability - No Disability (\textbf{ND}), Front Ankle broken (\textbf{FA}) and Back Ankle broken (\textbf{BA}). All the agents except `` GAN-MPC: $S^\mathbf{D} \subset S^\mathbf{I}$ '' are trained on the same set of demonstrations $\mathcal{X}^\mathbf{D}_s$. As described in Section \ref{sec:experiments}, `` GAN-MPC: $S^\mathbf{D} \subset S^\mathbf{I}$ '' is trained on $\mathcal{X}^\mathbf{D}_s$ but only a subset of the state variables are exposed.}
\label{fig:cheetah-run-experiments}
\end{figure}

We compare the performance of our proposed algorithm (\textbf{GAN-MPC}) with Behavioral Cloning (\textbf{BC}) and the Learnable-MPC formulation of \citet{xiao2022learning} that minimizes the L2 distance between the demonstrator and imitator trajectories. Unlike \citet{xiao2022learning}, we use a multi-layer perceptron instead of a Performer network \citep{choromanski2020rethinking} as a parameterized terminal cost model for minimalism. The terminal cost model of our GAN-MPC policy shares the same multi-layer perceptron structure. We compare two Learnable-MPC baselines: a) \textbf{L2-MPC-SA} that matches state-action trajectories and b) \textbf{L2-MPC-S} that matches state-only trajectories of the demonstrator and the imitator. In many practical applications, the entire state space of the demonstrator may not be observable or the state spaces of the demonstrator and the imitator may only overlap partially. GAN-MPC can be easily leveraged in such a situation by matching only the overlapping set of state variables. We study the case of partial observability in the \texttt{Cheetah-Run} environment. We train a set of GAN-MPC imitators on the demonstrator trajectories with all but one (velocity in forward direction) of the velocity state variables masked out -- see Table \ref{tab:cheetah-partial-state} in the supplementary material for details. We denote these agents by `` GAN-MPC: $S^\mathbf{D} \subset S^\mathbf{I}$ '' in Figure \ref{fig:cheetah-run-experiments}.

\begin{figure}
\centering
\begin{minipage}{.45\textwidth}
\includegraphics[width=0.7\textwidth]{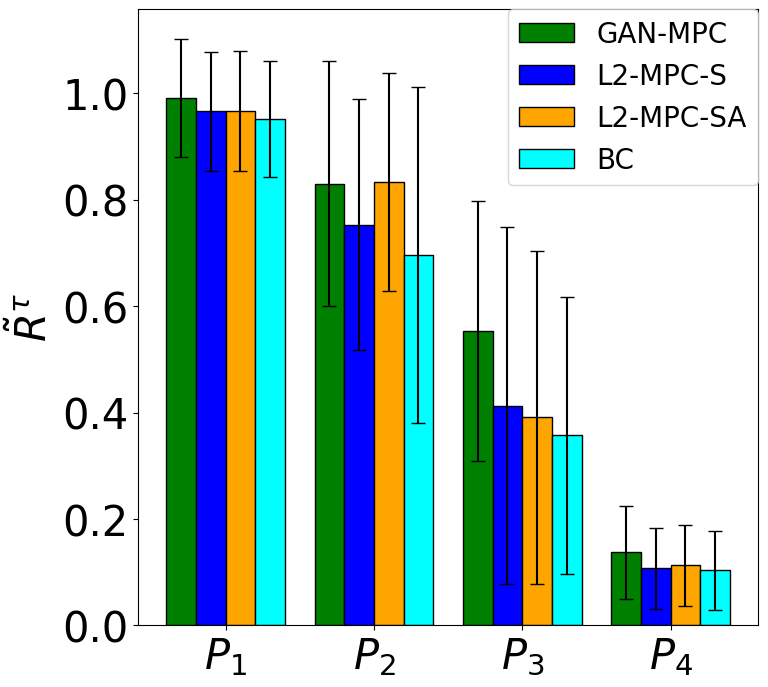}
\caption{Results of the \texttt{Pendulum-Swingup} experiment. The imitators are denoted by $P_x$ where $P$ stands for \emph{pole mass} and $x = P_{imitator}/P_{demonstrator}$.}
\label{fig:pendulum-results}
\end{minipage}
\hspace{0.2in}
\begin{minipage}{.45\textwidth}
\includegraphics[width=0.7\textwidth]{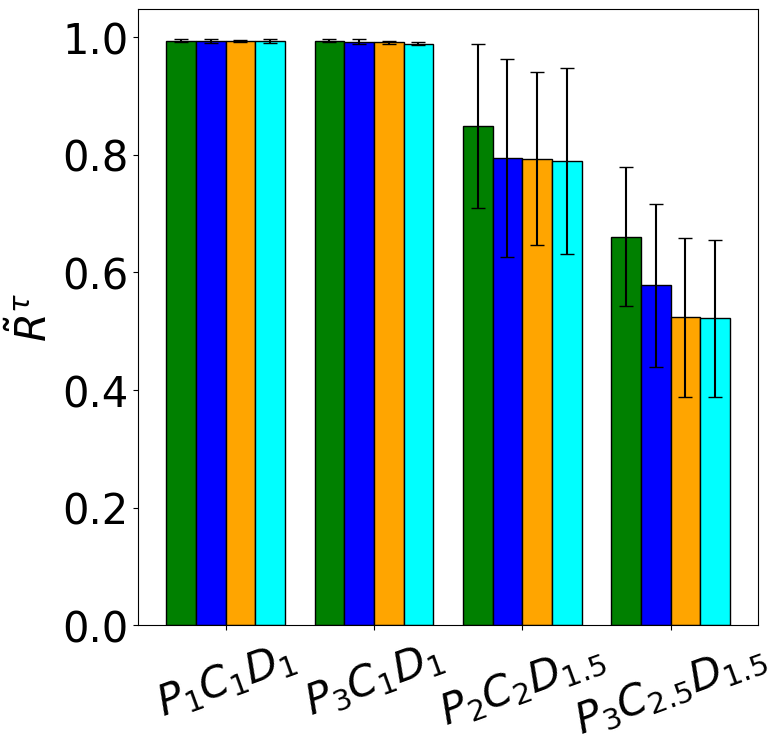}
\caption{Results of the \texttt{Cartpole-Balance} experiment. The imitators are denoted by $P_x C_y D_z$ where $P$, $C$ and $D$ stand for \emph{pole mass}, \emph{cart mass} and \emph{cart dimension}, respectively. The subscripts - $x$, $y$ and $z$ - denote ratios relative to the demonstrator, e.g. $x = P_{imitator}/P_{demonstrator}$. The legend of Figure \ref{fig:pendulum-results} has been followed.}
\label{fig:cartpole-results}
\end{minipage}%
\end{figure}

In all experiments, a training set of $50$ trajectories is collected from the demonstrator. \textbf{L2-MPC-SA}, \textbf{L2-MPC-S} and \textbf{GAN-MPC} imitators are allowed to interact with the environment for a total of $5000$ steps for \texttt{Cartpole-Balance} and \texttt{Pendulum-Swingup}; and $10000$ steps for \texttt{Cheetah-Run}. The performance of each agent is measured by rolling out $50$ trajectories with different random seeds and computing the average trajectory reward $R^\tau$. Figures \ref{fig:pendulum-results}, \ref{fig:cartpole-results} and \ref{fig:cheetah-run-experiments} provide a summary of the results. The bars represent means and the whiskers represent standard deviations. Tables \ref{tab:cartpole}, \ref{tab:cheetah} and \ref{tab:cheetah} in the supplementary material present the details. We measure the performance of the imitators in terms of average trajectory reward relative to the demonstrator, $\tilde{R}^\tau$ defined in Equation \ref{eq:evaluation-metric}. All hyperparameter values used in our experiments are presented in Table \ref{tab:hyperparameters} in the supplementary material.
\begin{equation}
    \tilde{R}^\tau=\frac{R^{\tau}_{imitator}}{R^{\tau}_{demonstrator}}
    \label{eq:evaluation-metric}
\end{equation}

\begin{figure}[t!]
    \centering
    \includegraphics[width=0.7\textwidth]{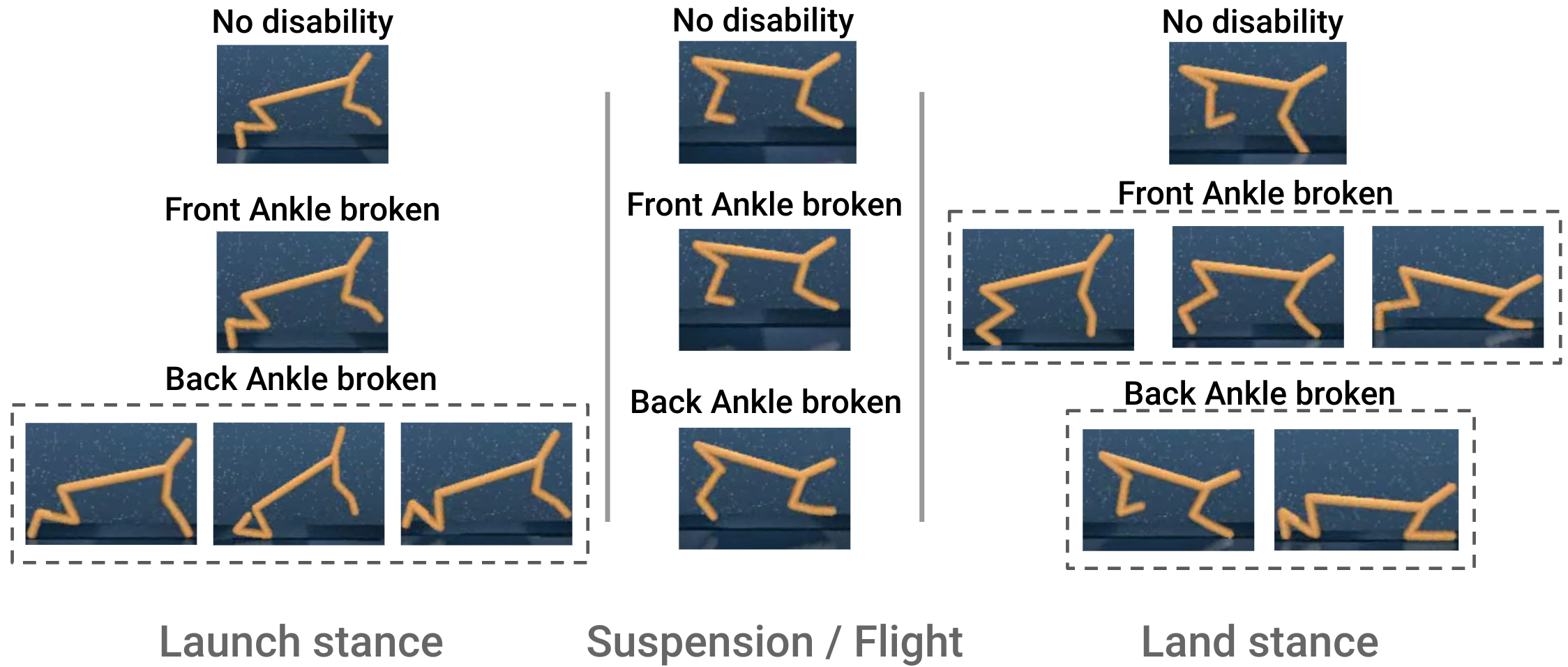}
    \caption{Characteristics of the galloping behavior learned by different imitators with different physical properties from the same set of demonstrations for the \texttt{Cheetah-Run} task. All the imitators have the same torso-mass ($2\times$ the demonstrator) but different types of disability, as marked in the figure. A cheetah's gallop consists of three phases: 1) ``Launch stance'', where the cheetah gathers propulsion to leap; 2) ``Suspension/Flight'', where the whole body of the cheetah is in the air; and 3) ``Land stance'', where the cheetah touches down in preparation for the next leap. The top row shows an imitator with no disability. It launches on the rear foot and lands on the front foot similar to the demonstrator which also does not have any disability. The middle row shows an imitator whose front ankle is broken. While it launches on the rear foot like the demonstrator, it learns that it can not land on the front foot since it would not be able to maintain stability due to the broken ankle. It learns to land with the rear foot down or both feet down or in a crouched position as viable alternatives. Finally, the bottom row shows an imitator whose back ankle is broken. While it often lands on the front foot like the demonstrator, it uses the front leg, back knee and sometimes the whole body for propulsion during launch. These results align with our goal of learning the demonstrator's behavior without having to copy their actions. Please visit our website for full videos of these behaviors.}
    \label{fig:cheetah-run-analysis}

\end{figure}

In Figures \ref{fig:pendulum-results}, \ref{fig:cartpole-results} and \ref{fig:cheetah-run-experiments}, we observe that GAN-MPC outperforms or matches the baselines in most of the settings. This validates the efficacy of our approach. We also observe that the performance of GAN-MPC gracefully degrades (like most of the baselines) as the dynamics of the imitator becomes more and more different from the demonstrator. In Figure \ref{fig:cheetah-run-analysis}, we demonstrate how the disabled imitators, in their quest to learn the fit demonstrator's skills, learn alternative strategies to work around their disabilities. This establishes GAN-MPC as a viable step towards achieving the goal of learning skills from non-identical experts without having to copy their actions. In Figure \ref{fig:cheetah-run-experiments}, we also observe that under partial observability of the demonstrator's state space the GAN-MPC agents (`` GAN-MPC: $S^\mathbf{D} \subset S^\mathbf{I}$ '') are able to learn the desired behavior and outperform the baselines that have access to the full state observations. This shows the viability of GAN-MPC as a method to learn skills from experts with non-identical dynamics and partial observability of their state spaces.
\vspace{-0.05in}
\section{Conclusions}
\label{sec:conclusion}
\vspace{-0.05in}
In this paper, we study imitation learning of MPC policies with parameterised cost functions. We consider the practical challenges of mismatch in the dynamics of the demonstrator and the imitator agents and partial observability of the state space of the demonstrator. We propose a novel approach called GAN-MPC that minimizes the statistical divergence between state-trajectories of the demonstator and the imitator using the GAN framework. Experiments on continuous control tasks of the DeepMind Control suite demonstrate the viability of the proposed method. The GAN-MPC framework needs significantly fewer samples of real world interaction of the imitator compared to RL based methods and this makes it viable for real world applications.
\vspace{-0.05in}
\section{Broader Societal Impacts and Limitations}
\label{sec:societal-impact}
\vspace{-0.05in}
Given the academic nature of the effort, we envisage no potential broader societal harm. If and when the work is explored for its utility in accessibility related use-cases, one will need a very rigorous analysis of failure modes to prevent any possible harm. The main limitations of our algorithm are a) it needs information about the actions as part of demonstrations, which may not be available or may be difficult to obtain, b) it makes a significant assumption that the action space must be the same and that there must be some overlap between the state spaces of the expert and the imitator. A more practical approach would be to relax these requirements, as it would allow us to use more diverse dataset from different robots performing the same tasks.

\section{Acknowledgements}
\label{sec:acknowledgement}
The authors would like to thank the authors of \citet{xiao2022learning} with special mention to Vikas Sindhwani, Tingnan Zhang, Anthony Francis and Sumeet Singh for helpful discussions and constructive criticism of our work.


\bibliographystyle{unsrtnat}
\bibliography{biblio}
\counterwithin{table}{section}
\renewcommand{\thetable}{A.\arabic{table}}
\section*{Supplementary Material}

\begin{table}[!h]
\scriptsize
\centering

\begin{minipage}{.7\textwidth}
\centering
\caption{Results of \texttt{CartPole-Balance} experiment. The demonstrator is the default agent from DeepMind Control \citep{tassa2018deepmind}. The imitators are denoted by $P_x C_y D_z$ where $P$, $C$ and $D$ stand for \emph{pole mass}, \emph{cart mass} and \emph{cart dimension} parameters, respectively. The subscripts - $x$, $y$ and $z$ - denote the value of the corresponding parameter relative to the demonstrator.}
\label{tab:cartpole}
\vspace{0.05in}
\begin{tabular}{l|c|c|c|c|}
\cline{2-5}
 & \multicolumn{4}{c|}{\textbf{Algorithms}} \\ \cline{2-5} 
 & \textbf{BC} & \textbf{L2-MPC-SA} & \textbf{L2-MPC-S} & \textbf{GAN-MPC} \\ \hline
\multicolumn{1}{|l|}{$P_1 C_1 D_1$} & $0.993\pm 0.003$ & $0.993\pm 0.002$ & $0.993\pm 0.003$ & $\mathbf{0.994\pm 0.002}$ \\ \hline
\multicolumn{1}{|l|}{$P_3 C_1 D_1$} & $0.989\pm 0.002$ & $0.991\pm 0.003$ & $0.992\pm 0.004$ & $\mathbf{0.994\pm 0.003}$ \\ \hline
\multicolumn{1}{|l|}{$P_2 C_2 D_{1.5}$} & $0.790\pm 0.158$ & $0.793\pm 0.147$ & $0.794\pm 0.168$ & $\mathbf{0.849\pm 0.140}$ \\ \hline
\multicolumn{1}{|l|}{$P_{3} C_{2.5} D_{1.5}$} & $0.522\pm 0.133$ & $0.524\pm 0.135$ & $0.578\pm 0.139$ & $\mathbf{0.661\pm 0.118}$ \\ \hline
\end{tabular}
\end{minipage}%
\hfill
\begin{minipage}{.7\textwidth}
\centering
\caption{Results of \texttt{Pendulum-Swingup} experiment. The demonstrator is the default agent from DeepMind Control \citep{tassa2018deepmind}. The imitators are denoted by $P_x$ where $P$ stands for \emph{pole mass}. The subscript $x = \frac{P_{imitator}}{P_{demonstrator}}$.}
\label{tab:pendulum}
\vspace{0.05in}
\begin{tabular}{l|c|c|c|c|}
\cline{2-5}
 & \multicolumn{4}{c|}{\textbf{Algorithms}} \\ \cline{2-5} 
 & \multicolumn{1}{c|}{\textbf{BC}} & \multicolumn{1}{c|}{\textbf{L2-MPC-SA}} & \multicolumn{1}{c|}{\textbf{L2-MPC-S}} & \multicolumn{1}{c|}{\textbf{GAN-MPC}} \\ \hline
\multicolumn{1}{|l|}{$P_1$} & $0.951\pm 0.109$ & $0.967\pm 0.113$ & $0.966\pm 0.111$ & $\mathbf{0.992\pm 0.111}$ \\ \hline
\multicolumn{1}{|l|}{$P_2$} & $0.696\pm 0.316$ & $\mathbf{0.833\pm 0.205}$ & $0.753\pm 0.236$ & $0.830\pm 0.230$ \\ \hline
\multicolumn{1}{|l|}{$P_3$} & $0.357\pm 0.260$ & $0.391\pm 0.313$ & $0.413\pm 0.336$ & $\mathbf{0.553\pm 0.245}$ \\ \hline
\multicolumn{1}{|l|}{$P_4$} & $0.103\pm 0.074$ & $0.113\pm 0.076$ & $0.107\pm 0.076$ & $\mathbf{0.137 \pm 0.088}$ \\ \hline
\end{tabular}

\end{minipage}

\end{table}

\begin{table}[]
\scriptsize
\centering
\caption{Results of \texttt{Cheetah-Run} experiment. The demonstrator is the default agent from DeepMind Control \citep{tassa2018deepmind}. We evaluate our algorithm for $12$ different imitators obtained by changing the torso-mass $M_x$, where $x = \frac{M_{imitator}}{M_{demonstrator}}$, and simulating disability. The columns marked $S^\mathbf{D}\subset S^\mathbf{I}$ shows results for imitation learning from a demonstrator whose state-space is partially observable.}
\label{tab:cheetah}

\begin{subtable}{.7\linewidth}
\centering
\caption{$M_{imitator} = M_{demonstrator}$}
\label{tab:same-mass-cheetah}

\begin{tabular}{l|ccccc|}
\cline{2-6}
 & \multicolumn{5}{c|}{\textbf{Algorithms}} \\ \cline{2-6} 
 & \multicolumn{4}{c|}{$S^\mathbf{D}=S^\mathbf{I}$} & \multicolumn{1}{c|}{$S^\mathbf{D}\subset S^\mathbf{I}$} \\ \cline{2-6} 
 & \multicolumn{1}{c|}{\textbf{BC}} & \multicolumn{1}{c|}{\textbf{L2-MPC-SA}} & \multicolumn{1}{c|}{\textbf{L2-MPC-S}} & \multicolumn{1}{c|}{\textbf{GAN-MPC}} & \multicolumn{1}{c|}{\textbf{GAN-MPC}} \\
\hline
\multicolumn{1}{|l|}{$F_1 B_1$} & \multicolumn{1}{c|}{$0.046\pm 0.026$} & \multicolumn{1}{c|}{$0.088\pm 0.052$} & \multicolumn{1}{c|}{$0.089\pm 0.046$} & \multicolumn{1}{c|}{$\mathbf{0.141\pm 0.055}$} & \multicolumn{1}{c|}{$0.134\pm 0.055$} \\ \hline
\multicolumn{1}{|l|}{$F_0 B_1$} & \multicolumn{1}{c|}{$0.053\pm 0.014$} & \multicolumn{1}{c|}{$0.083\pm 0.059$} & \multicolumn{1}{c|}{$0.083\pm 0.057$} & \multicolumn{1}{c|}{$\mathbf{0.096\pm 0.044}$} & \multicolumn{1}{c|}{$0.091\pm 0.049$}\\ \hline
\multicolumn{1}{|l|}{$F_1 B_0$} & \multicolumn{1}{c|}{$0.049\pm 0.021$} & \multicolumn{1}{c|}{$0.029 \pm 0.030$} & \multicolumn{1}{c|}{$0.027\pm 0.021$} & \multicolumn{1}{c|}{$\mathbf{0.077\pm 0.026}$} & \multicolumn{1}{c|}{$0.076\pm 0.022$}\\ \hline
\end{tabular}
\end{subtable}%
\hfill
\begin{subtable}{.7\linewidth}
\centering
\caption{$M_{imitator} = 1.25\times M_{demonstrator}$}
\label{tab:same-mass-cheetah}

\begin{tabular}{l|ccccc|}
\cline{2-6}
 & \multicolumn{5}{c|}{\textbf{Algorithms}} \\ \cline{2-6} 
 & \multicolumn{4}{c|}{$S^\mathbf{D}=S^\mathbf{I}$} & \multicolumn{1}{c|}{$S^\mathbf{D}\subset S^\mathbf{I}$} \\ \cline{2-6} 
 & \multicolumn{1}{c|}{\textbf{BC}} & \multicolumn{1}{c|}{\textbf{L2-MPC-SA}} & \multicolumn{1}{c|}{\textbf{L2-MPC-S}} & \multicolumn{1}{c|}{\textbf{GAN-MPC}} & \multicolumn{1}{c|}{\textbf{GAN-MPC}} \\
\hline
\multicolumn{1}{|l|}{$F_1 B_1$} & \multicolumn{1}{c|}{$0.039\pm 0.017$} & \multicolumn{1}{c|}{$0.109\pm 0.046$} & \multicolumn{1}{c|}{$0.115\pm 0.055$} & \multicolumn{1}{c|}{$\mathbf{0.153\pm 0.059}$} & \multicolumn{1}{c|}{$0.139\pm 0.053$} \\ \hline
\multicolumn{1}{|l|}{$F_0 B_1$} & \multicolumn{1}{c|}{$0.051\pm 0.026$} & \multicolumn{1}{c|}{$0.095\pm 0.054$} & \multicolumn{1}{c|}{$0.087\pm 0.060$} & \multicolumn{1}{c|}{$\mathbf{0.115\pm 0.055}$} & \multicolumn{1}{c|}{$0.106\pm 0.052$}\\ \hline
\multicolumn{1}{|l|}{$F_1 B_0$} & \multicolumn{1}{c|}{$0.048\pm 0.016$} & \multicolumn{1}{c|}{$0.030 \pm 0.020$} & \multicolumn{1}{c|}{$0.032\pm 0.025$} & \multicolumn{1}{c|}{$\mathbf{0.088\pm 0.017}$} & \multicolumn{1}{c|}{$0.087\pm 0.018$}\\ \hline
\end{tabular}
\end{subtable}%
\hfill
\begin{subtable}{.7\linewidth}
\centering
\caption{$M_{imitator} = 1.5\times M_{demonstrator}$}
\label{tab:same-mass-cheetah}

\begin{tabular}{l|ccccc|}
\cline{2-6}
 & \multicolumn{5}{c|}{\textbf{Algorithms}} \\ \cline{2-6} 
 & \multicolumn{4}{c|}{$S^\mathbf{D}=S^\mathbf{I}$} & \multicolumn{1}{c|}{$S^\mathbf{D}\subset S^\mathbf{I}$} \\ \cline{2-6} 
 & \multicolumn{1}{c|}{\textbf{BC}} & \multicolumn{1}{c|}{\textbf{L2-MPC-SA}} & \multicolumn{1}{c|}{\textbf{L2-MPC-S}} & \multicolumn{1}{c|}{\textbf{GAN-MPC}} & \multicolumn{1}{c|}{\textbf{GAN-MPC}} \\
\hline
\multicolumn{1}{|l|}{$F_1 B_1$} & \multicolumn{1}{c|}{$0.073\pm 0.020$} & \multicolumn{1}{c|}{$0.107\pm 0.055$} & \multicolumn{1}{c|}{$0.119\pm 0.073$} & \multicolumn{1}{c|}{$\mathbf{0.151\pm 0.059}$} & \multicolumn{1}{c|}{$0.140\pm 0.061$} \\ \hline
\multicolumn{1}{|l|}{$F_0 B_1$} & \multicolumn{1}{c|}{$0.068\pm 0.017$} & \multicolumn{1}{c|}{$0.103\pm 0.061$} & \multicolumn{1}{c|}{$0.105\pm 0.056$} & \multicolumn{1}{c|}{$\mathbf{0.129\pm 0.044}$} & \multicolumn{1}{c|}{$0.121\pm 0.049$}\\ \hline
\multicolumn{1}{|l|}{$F_1 B_0$} & \multicolumn{1}{c|}{$0.038\pm 0.023$} & \multicolumn{1}{c|}{$0.035 \pm 0.022$} & \multicolumn{1}{c|}{$0.034\pm 0.025$} & \multicolumn{1}{c|}{$\mathbf{0.091\pm 0.019}$} & \multicolumn{1}{c|}{$0.088\pm 0.018$}\\ \hline
\end{tabular}
\end{subtable}%
\hfill

\begin{subtable}{.7\linewidth}
\centering
\caption{$M_{imitator} = 2\times M_{demonstrator}$}
\label{tab:same-mass-cheetah}

\begin{tabular}{l|ccccc|}
\cline{2-6}
 & \multicolumn{5}{c|}{\textbf{Algorithms}} \\ \cline{2-6} 
 & \multicolumn{4}{c|}{$S^\mathbf{D}=S^\mathbf{I}$} & \multicolumn{1}{c|}{$S^\mathbf{D}\subset S^\mathbf{I}$} \\ \cline{2-6} 
 & \multicolumn{1}{c|}{\textbf{BC}} & \multicolumn{1}{c|}{\textbf{L2-MPC-SA}} & \multicolumn{1}{c|}{\textbf{L2-MPC-S}} & \multicolumn{1}{c|}{\textbf{GAN-MPC}} & \multicolumn{1}{c|}{\textbf{GAN-MPC}} \\
\hline
\multicolumn{1}{|l|}{$F_1 B_1$} & \multicolumn{1}{c|}{$0.078\pm 0.038$} & \multicolumn{1}{c|}{$0.119\pm 0.062$} & \multicolumn{1}{c|}{$0.135\pm 0.066$} & \multicolumn{1}{c|}{$\mathbf{0.159\pm 0.055}$} & \multicolumn{1}{c|}{$0.151\pm 0.057$} \\ \hline
\multicolumn{1}{|l|}{$F_0 B_1$} & \multicolumn{1}{c|}{$0.079\pm 0.028$} & \multicolumn{1}{c|}{$0.108\pm 0.049$} & \multicolumn{1}{c|}{$0.111\pm 0.059$} & \multicolumn{1}{c|}{$\mathbf{0.136\pm 0.055}$} & \multicolumn{1}{c|}{$0.130\pm 0.059$}\\ \hline
\multicolumn{1}{|l|}{$F_0 B_1$} & \multicolumn{1}{c|}{$0.056\pm 0.016$} & \multicolumn{1}{c|}{$0.043 \pm 0.026$} & \multicolumn{1}{c|}{$0.044\pm 0.027$} & \multicolumn{1}{c|}{$\mathbf{0.105\pm 0.021}$} & \multicolumn{1}{c|}{$0.100\pm 0.020$}\\ \hline
\end{tabular}
\end{subtable}%

\end{table}


\begin{table}[]
    \caption{Table of hyperparameters}
    \label{tab:hyperparameters}
    \centering
\begin{tabular}{|c|ccc|}
\hline
\textbf{Environment}                            & \multicolumn{1}{c|}{\ \ \textbf{Cartpole-Balance}\ \ } & \multicolumn{1}{c|}{\ \ \textbf{Pendulum-Swingup}\ \ } & \ \ \textbf{Cheetah-Run}\ \  \\ \hline
Trajectory Maximum Length                       & \multicolumn{2}{c|}{500}                                                                       & 500                  \\ \hline
Optimizer                                       & \multicolumn{3}{c|}{Adam Optimizer}                                                                                    \\ \hline
Dynamics Model Network                          & \multicolumn{3}{c|}{4 layers, 200 hidden neurons, ReLU}                                                                \\ \hline
Cost Model Network                              & \multicolumn{3}{c|}{4 layers, 128 hidden neurons, ReLU}                                                                \\ \hline
Number of expert demonstrations                 & \multicolumn{3}{c|}{50}                                                                                                \\ \hline
$N^{pre}$ (pre-training dynamics model)                 & \multicolumn{3}{c|}{2}                                                                                                 \\ \hline
K (number of env rollouts)                      & \multicolumn{3}{c|}{1}                                                                                                 \\ \hline
$N^{dyn}$ (update dynamics model during training)       & \multicolumn{2}{c|}{2}                                                                          & 5                    \\ \hline
$\gamma$ (terminal cost weight)                          & \multicolumn{3}{c|}{1}                                                                                                 \\ \hline
$N^{MPC}$                                               & \multicolumn{2}{c|}{10}                                                                         & 20                   \\ \hline
H (MPC Horizon )                                & \multicolumn{3}{c|}{10}                                                                                                \\ \hline
P (generator rollout for predicting state seq.) & \multicolumn{3}{c|}{10}                                                                                                \\ \hline
B (Batch Size)                                  & \multicolumn{3}{c|}{128}                                                                                               \\ \hline
Learning Rate                                   & \multicolumn{3}{c|}{1e-5}                                                                                              \\ \hline
\end{tabular}
\end{table}

    

\begin{table}[]
\centering
\caption{Demonstrator rewards}
\label{tab:expert-rewards}
\begin{tabular}{|c|ccc|}
\hline
\textbf{Environment}              & \multicolumn{1}{c|}{\ \ \textbf{Cartpole-Balance}\ \ } & \multicolumn{1}{c|}{\ \ \textbf{Pendulum-Swingup}\ \ } & \ \ \textbf{Cheetah-Run}\ \  \\ \hline
\ Algorithm to generate expert demo\  & \multicolumn{3}{c|}{SAC}                                                                                               \\ \hline
Expert reward till 1000 timesteps & \multicolumn{1}{c|}{997}                       & \multicolumn{1}{c|}{880}                       & 950                  \\ \hline
\end{tabular}
\end{table}


\def\checkmark{\tikz\fill[scale=0.4](0,.35) -- (.25,0) -- (1,.7) -- (.25,.15) -- cycle;} 

\begin{table}[]
\centering
\caption{State spaces of the demonstrator $\mathbf{D}$ and the imitator $\mathbf{I}$ in our experiment on \texttt{Cheetah-Run} where the state space of the demonstrator is partially observable to the imitator. The imitator can observe all the position state-variables of the demonstrator. All the velocity state-variables of the demonstrator are masked from the imitator except \texttt{rootx}.}
\label{tab:cheetah-partial-state}


\begin{tabular}{|c|c|cc|cc|}
\hline
\multirow{2}{*}{\textbf{State variable}} & \multirow{2}{*}{\ \ \textbf{Symbol in DM-Control}\ \ } & \multicolumn{2}{c|}{\textbf{position}}                                                                                                                & \multicolumn{2}{c|}{\textbf{velocity}}                                                                                                                 \\ \cline{3-6} 
                                               &                                      & \multicolumn{1}{c|}{\begin{tabular}[c]{@{}c@{}}\ \ $\mathbf{D}$\end{tabular}} & \begin{tabular}[c]{@{}c@{}}\ \ $\mathbf{I}$\end{tabular} & \multicolumn{1}{c|}{\begin{tabular}[c]{@{}c@{}}\ \ $\mathbf{D}$\end{tabular}} & \begin{tabular}[c]{@{}c@{}}\ \ $\mathbf{I}$\end{tabular} \\ \hline
x-coordinate of front tip                      & \texttt{rootx}                                & \multicolumn{1}{c|}{$\times$}                                                             & $\times$                                                              & \multicolumn{1}{c|}{\checkmark}                                                             & \checkmark                                                             \\ \hline
z-coordinate of front tip                      & \texttt{rootz}                                & \multicolumn{1}{c|}{\checkmark}                                                            & \checkmark                                                             & \multicolumn{1}{c|}{\checkmark}                                                             & $\times$                                                              \\ \hline
angle of front tip                             & \texttt{rooty}                                & \multicolumn{1}{c|}{\checkmark}                                                            & \checkmark                                                             & \multicolumn{1}{c|}{\checkmark}                                                             & $\times$                                                              \\ \hline
angle of back thigh                            & \texttt{bthigh}                               & \multicolumn{1}{c|}{\checkmark}                                                            & \checkmark                                                             & \multicolumn{1}{c|}{\checkmark}                                                             & $\times$                                                              \\ \hline
angle of back second rotor                     & \texttt{bshin}                                & \multicolumn{1}{c|}{\checkmark}                                                            & \checkmark                                                             & \multicolumn{1}{c|}{\checkmark}                                                             & $\times$                                                              \\ \hline
angle of back foot                             & \texttt{bfoot}                                & \multicolumn{1}{c|}{\checkmark}                                                            & \checkmark                                                             & \multicolumn{1}{c|}{\checkmark}                                                             & $\times$                                                              \\ \hline
angle of front thigh                           & \texttt{fthigh}                               & \multicolumn{1}{c|}{\checkmark}                                                            & \checkmark                                                             & \multicolumn{1}{c|}{\checkmark}                                                             & $\times$                                                              \\ \hline
angle of front second rotor                    & \texttt{fshin}                                & \multicolumn{1}{c|}{\checkmark}                                                            & \checkmark                                                             & \multicolumn{1}{c|}{\checkmark}                                                             & $\times$                                                              \\ \hline
angle of front foot                            & \texttt{ffoot}                                & \multicolumn{1}{c|}{\checkmark}                                                            & \checkmark                                                             & \multicolumn{1}{c|}{\checkmark}                                                             & $\times$                                                              \\ \hline
\end{tabular}

\end{table}

\end{document}